\documentclass[conference]{IEEEtran}
\IEEEoverridecommandlockouts
\usepackage{subfigure}
\usepackage{amsmath,amssymb,amsfonts}
\usepackage{cleveref}
\usepackage{bm}
\usepackage{multirow}
\usepackage{algorithmic}
\usepackage{graphicx}
\usepackage{textcomp}
\usepackage{xcolor}
\usepackage{comment}

\def\BibTeX{{\rm B\kern-.05em{\sc i\kern-.025em b}\kern-.08em
    T\kern-.1667em\lower.7ex\hbox{E}\kern-.125emX}}
\begin{document}

\title{3D-FlowNet: Event-based optical flow estimation with 3D representation\\

}

\author{\IEEEauthorblockN{Haixin SUN\IEEEauthorrefmark{1},
 Minh-Quan DAO\IEEEauthorrefmark{2},
Vincent FREMONT\IEEEauthorrefmark{3}}
\IEEEauthorblockA{Nantes Universit\'{e}, Ecole Centrale de Nantes, CNRS LS2N\\
Nantes, France\\
Email: \IEEEauthorrefmark{1}haixin.sun@ec-nantes.fr,
\IEEEauthorrefmark{2}minh-quan.dao@ec-nantes.fr,
\IEEEauthorrefmark{3}vincent.fremont@ec-nantes.fr}}

\maketitle

\begin{abstract}

Event-based cameras can overpass frame-based cameras limitations for important tasks such as high-speed motion detection during self-driving cars navigation in low illumination conditions. The event cameras' high temporal resolution and high dynamic range, allow them to work in fast motion and extreme light scenarios. However, conventional computer vision methods, such as Deep Neural Networks, are not well adapted to work with event data as they are asynchronous and discrete. Moreover, the traditional 2D-encoding representation methods for event data, sacrifice the time resolution. In this paper, we first improve the 2D-encoding representation by expanding it into three dimensions to better preserve the temporal distribution of the events. We then propose 3D-FlowNet, a novel network architecture that can process the 3D input representation and output optical flow estimations according to the new encoding methods. A self-supervised training strategy is adopted to compensate the lack of labeled datasets for the event-based camera. Finally, the proposed network is trained and evaluated with the Multi-Vehicle Stereo Event Camera (MVSEC) dataset. The results show that our 3D-FlowNet outperforms state-of-the-art approaches with less training epoch (30 compared to 100 of Spike-FlowNet).

\end{abstract}


\section{Introduction}

An Autonomous Vehicle (AV) requires an accurate perception of its surrounding environment to reliably and safely operate. The perception system of an AV can transform raw sensory data into semantic information \cite{8621614}, and frame-based monocular cameras are one of the most commonly used sensors for this purpose. They synchronously transmit raw images, frame by frame, at a fixed rate. This feature as the major drawbacks of low temporal resolution, redundant information and low dynamic range. Few years ago, event-based cameras, a bio-inspired technology of silicon retinas, have been proposed to overcome those limitations and to solve both classical and new computer vision tasks \cite{4444573,6889103}. An event-based camera can have a dynamic range of 130 \(dB\) and a minimum of 3 \(\mu s\) latency. Those advantages allow the event-based camera to work in extreme scenarios with low light conditions and fast motions. Typically, event-based cameras are used as sensing modalities on Unmanned aerial vehicle (UAV) \cite{evdodge}, mobile robots \cite{10.3389/fnins.2013.00223} or wearable electronics \cite{7599576}, where operations are under unrealistic lighting conditions and sensitive to the temporal resolution. The main applications for event-based cameras are object stracking \cite{10.3389/fnins.2013.00223}, surveillance and monitoring \cite{1706816}, and optical flow estimation \cite{DAViS-Camera-Optical-Flow,Zhu-RSS-18}. Nowadays, more and more researchers focus on using the event-based cameras for autonomous driving. \cite{Maqueda2018} proposed a method that can predict the vehicle's steering angle according to the event data, and \cite{ddd20} proposed a dataset that contains event data along with the vehicle state and CAN-bus data.
    \begin{figure}[!t]
      \centering
        \includegraphics[width=0.8\linewidth]{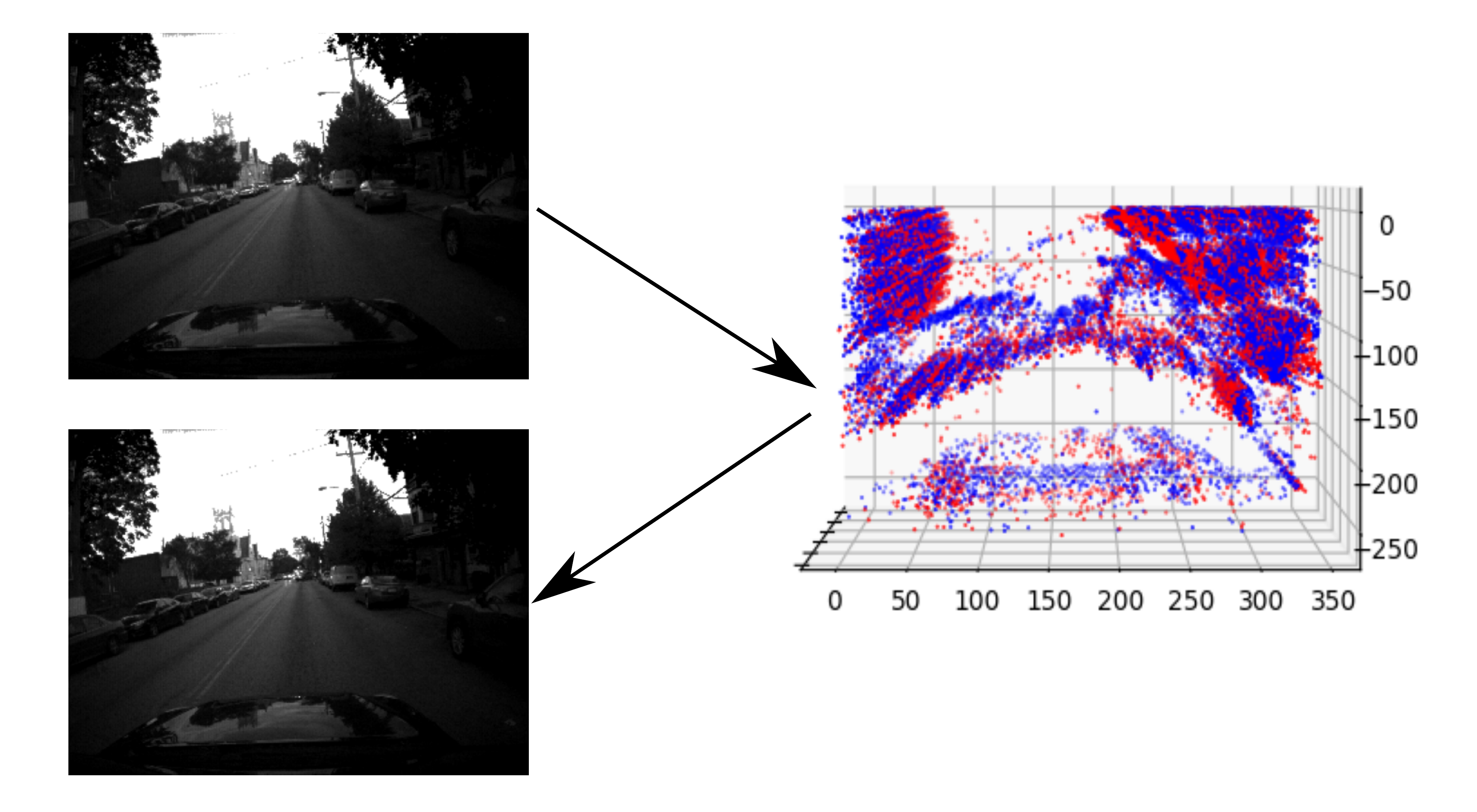}
        \includegraphics[width=0.8\linewidth]{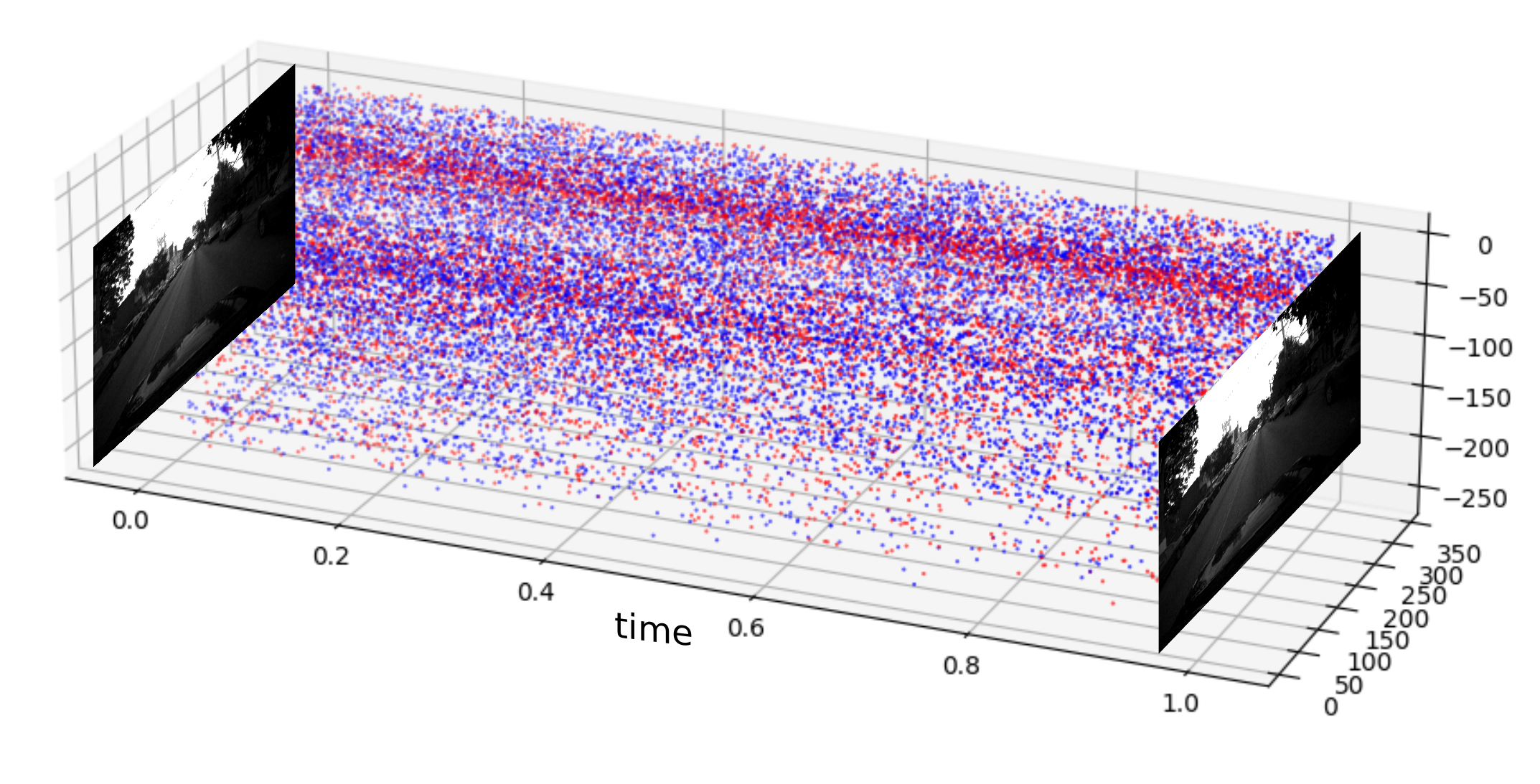}
      \caption{Visualization of the event data between two gray scale image.}
      \label{fig:events-Visualization}
    \end{figure}

Event-based cameras are asynchronous devices that detect changes in log brightness intensity. When the variation of the brightness of a pixel reaches the threshold, the camera generates an event. The event is usually in the format of a tuple, \( e = (x, y, t, p)\), where \((x, y)\) is the pixel's position, \(t\) is the precise timestamp of the event which is accurate up to microseconds, and the polarity \(p\) of the change that indicates whether the pixel became brighter or darker. Fig. \ref{fig:events-Visualization} shows the visualization of the event data between two frame-based gray-scale images. The positive events are shown in red, and the negative events are in blue. Between two consecutive images, there is a quasi-continuous stream of events that represents all the brightness change between the two images. The event-based camera's asynchronous nature and tracking in the log image space offer several advantages over traditional frame-based cameras, including extremely low latency for detecting high-speed objects, a very high dynamic range for the poor light conditions, and significantly lower power consumption.

The cameras' unique output, on the other hand, presents new challenges in algorithm developments. Indeed, the events are transmitted asynchronously and lacks the pixel's absolute value and spatial neighborhood. The algorithms for traditional frame images such as optical flow or object detection are no longer valid. As a result, a significant research effort has been made to develop new algorithms for event-based cameras to solve these traditional vision problems.

Within Deep Learning area, there exist several works that train a neural networks to estimate the optical flow in a self-supervised manner. Zhu et al. \cite{Zhu-RSS-18} accumulate the events into the image-like frames and calculate the optical flow using an encoder-decoder network. Their encoding method loses the temporal information because they summarize the events stream into a four-channel image. Lee et al. \cite{Spike-FlowNet} try to solve this problem by proposing a deep hybrid neural network architecture called Spike-FlowNet. The use of the Spiking Neural Network allows the approach to process the data asynchronously. So it can best preserve the properties of the event data. However, the training of the Spiking Neural Network is quite slow and unstable. So, although the neural networks avoid the complex problem of modeling and algorithm developments, the encoding representation for the event data and the neural network's design still need to be improved. 

The main contribution of this paper is to propose a new encoding method and the corresponding neural network architecture to process an event data stream. We proposed a 3D encoding representation that can better preserve the temporal nature of the event data. We also present the 3D-FlowNet, a novel neural network architecture that can process the 3D input and generate optical flow estimations. Finally, We train and evaluate the proposed 3D-FlowNet using the Multi-Vehicle Stereo Event Camera (MVSEC) dataset \cite{MVSEC}. The results show that our approach outperforms current state-of-the-art methods, we achieve \(13\% \) improvement compared to the Spike-FlowNet\cite{Spike-FlowNet}, and \(32\%\) compared to the EV-FlowNet\cite{Zhu-RSS-18}. 

The paper is structured as follows: In Section II, we discuss the related work. In section III, we present the methodology, covering the encoding method for the event data and the corresponding neural network architecture. This section also discusses the self-supervised training strategy. In section IV, we present the experimental results, including training details and the evaluation metrics. We also discuss the comparison results with state-of-the-art approaches.

\section{Related Work}
    
Due to the properties of the event-based camera, there has been a lot of interest in developing algorithms that take advantage of them, and optical flow estimation is one of the addressed topics. Benosman et al. \cite{Event-Based-Visual-Flow} fit a plane to the events in spatial-temporal spaces and then estimate the optical flow. Bardow et al. \cite{Simultaneous-Optical-Flow-and-Intensity-Estimation} formulate the flow estimation as a convex optimization
problem that solves for the image intensity and flow jointly. Almatrafi et al. \cite{DAViS-Camera-Optical-Flow} calculate the spatial and temporal gradients on the frame image and events data, respectively, and then estimate the optical flow by solving the classical optical flow equation.  

Besides the traditional optical flow algorithms for the event camera, there are also several model-free methods that use a deep neural networks to predict the optical flow. Zhu et al. \cite{Zhu-RSS-18} accumulate the event into the image like frames and use an encoder-decoder network architecture to estimate the optical flow. The event data are then encoded into a four-channel image representing: Positive events counting, negative events counting, latest timestamp of positive events, and latest timestamp of negative events. This encoding method loses the temporal information because the older timestamp are filtered out. Lee et al. \cite{Spike-FlowNet} try to solve this problem by proposing a deep hybrid neural network architecture called Spike-FlowNet, a hybrid structure between regular Neural Networks (NN) and Spiking Neural Networks (SNN). Due to the use of the SNN, the events are processed asynchronously to preserve the temporal information of the event data. However, the training of the Spike-FlowNet is relatively slow and unstable. Because the activation function of SNN is not continuous, the backpropagation algorithm can not be directly used to train the SNN.

For the networks' training, several works focuses on self-supervised training for the optical flow prediction because of the lack of labeled event-based datasets. Yu et al. \cite{jjyu2016unsupflow} proposed a network that can learn optical flow from brightness constancy and motion smoothness. Based on that, Meister et al. \cite{UnFlow} improve the quality of the flow by applying a bidirectional census loss to achieve better performance with less training time. \cite{Zhu-RSS-18,Spike-FlowNet} adopt this self-supervised strategy for event-based camera and achieve similar performances.

\section{Proposed Approach}
    \begin{figure*}[!ht]
      \centering
        \subfigure[Example of an event image and a gray scale image]{
            \includegraphics[width=0.4\linewidth]{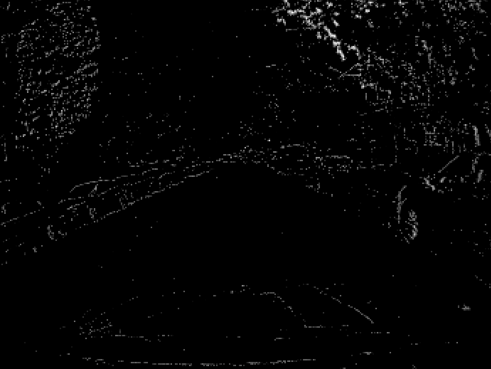}
            \includegraphics[width=0.4\linewidth]{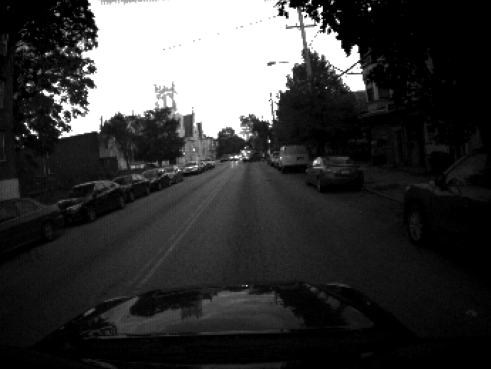}
        }
        \subfigure[four channels for event data]{
            \includegraphics[width=0.8\linewidth]{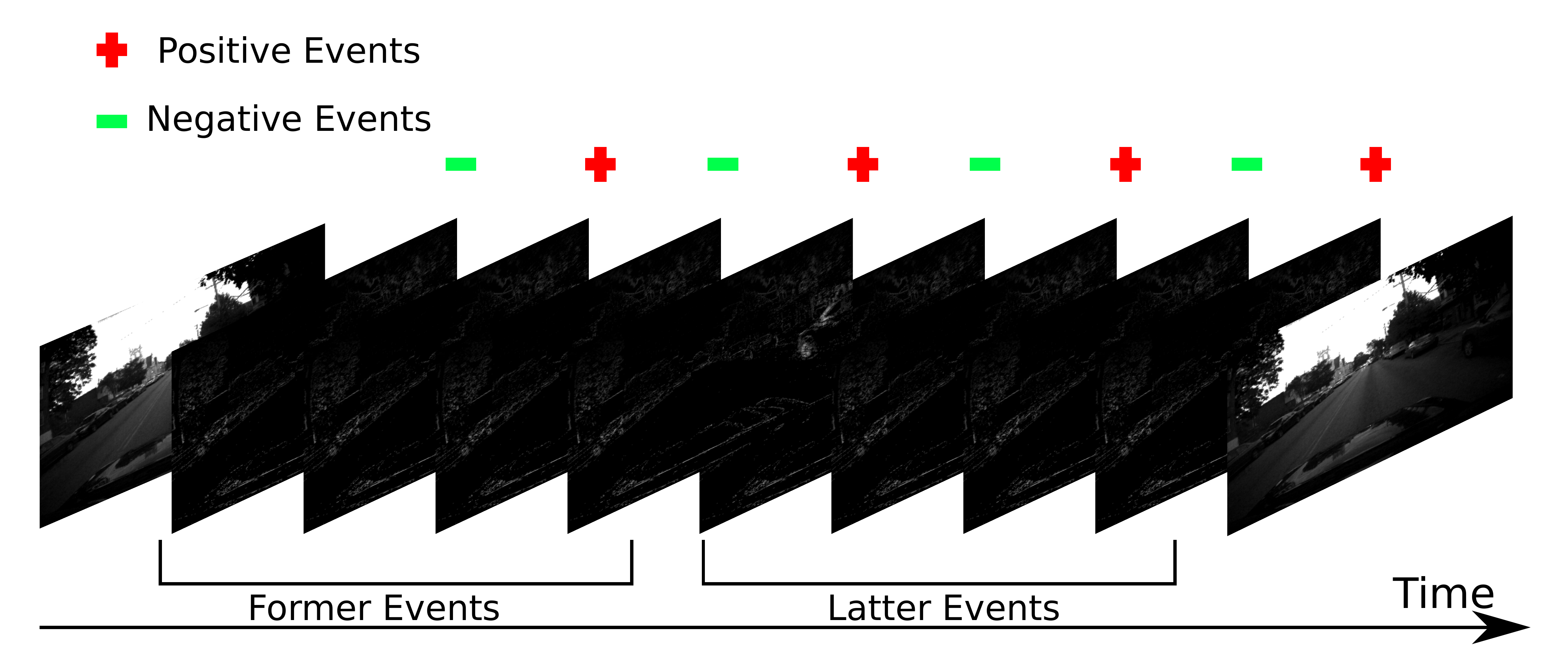}
        }
      \caption{Visualization of our event encoding representation. (a) is one slice of the event image \(I_{slice} = (1,1,H,W)\), and the brighter represents the more recent timestamp value. (b) is an example of the event representation where \(D = 8\). }
      \label{fig:input}
    \end{figure*}
\par In this section, we explain our approach in details. In \ref{sec:event-encoding}, we describe our event encoding method, which encodes a group of event measurements into an 3D temporal-spatial event image. In \ref{sec:Network}, we describe the architecture of our network, which uses the 3D convolutions to process the spatial-temporal measurements and output the pixel-wise optical flow. Finally, in \ref{sec:self-supervised-loss}, we describe the training strategy and the self-supervised loss is also discussed.

\subsection{Event Data Encoding Method}
\label{sec:event-encoding}
\par The event-based camera records the log intensity change of each pixel of the artificial retina, and generates an event whenever the log intensity changes over the threshold \(\theta\):
    \begin{equation}
         log(I_{t+1}) - log(I_{t}) \geq \theta
          \label{eqevlog}
    \end{equation}
The event measurement is in the format of tuple which consists of location of the pixel, timestamp of the event and polarity of the change:
    \begin{equation}
        e = (x,y,t,p)
          \label{eqevmes}
    \end{equation}
    
Because the events are transmitted asynchronously, they cannot be immediately fed into standard convolutional neural network layers. It is therefore important to keep the necessary information while generate the encoding representation from the event stream.  

Several prior works have proposed different methods that transform the event output into a synchronous image-like representation. In EV-FlowNet \cite{Zhu-RSS-18}, only the latest pixel-wise timestamps and the event counts are used to encode the event representation. However, fast motions and dense scenarios can enormously overlap per-pixel timestamp information. In \cite{zhu35,howto}, the time domain is discretized to preserve the temporal distributions. To improve the resolution and the temporal domain beyond the number of bins, the authors insert events into this volume using a linearly weighted accumulation similar to bilinear interpolation. However, the number of input channels increases significantly as the time dimensions are finely discretized, further increasing the computation time for encoding and forward propagation.

Considering all the methods discussed before, we propose in this work, a novel input representation that can better exploit the information in the event data with less computation complexity. Given a set of \(N\) input events \(\mathit{E}_N = {(x_i, y_i, t_i, p_i)}, i \in [1,N]\), and a time depth \(D\) to discretize the time dimension of event data, we accumulate each group of event into images as follows: 
    \begin{equation}\label{eqevimage}
        \begin{split}
            t_{norm} &= (t-t_0)/(t_N-t_1)*(D-1) \\
            I(x,y,t,p) &= \sum_{i} \delta(p-p_i)k_b(x-x_i)k_b(y-y_i)k_b(t-t_{norm})\\
            k_b(a) &= max(0,1-\lvert a\rvert)
        \end{split}
    \end{equation}
Here, \((x,y)\) denotes the position of the event, \(p\) is the polarity of the event, and \(\delta\) is the Kronecker delta operator. \(k_b(\cdot)\) denotes bi-linear sampling kernel. The generated event image \(I\) is a \((2,D,H,W)\) matrix, where the number 2 represents the positive and negative polarity, D is the discretized time depth, and \((H,W)\) are respectively the height and width of the image. Then we split the event image into former and latter groups through the time dimension and obtained a new event image with the shape of \((4,\frac{D}{2},H,W)\). Here the number 4 represents the four channels: Former positive events, former negative events, latter positive events, latter negative events. Fig. \ref{fig:input} shows the proposed input representation. Fig. \ref{fig:input}. (a) is the visualization of the event image and the relative grayscale image, left is one slice of the event image, and the brighter represents the more recent timestamp value. Fig. \ref{fig:input}. (b) is an example of the event representation where \(D = 8\).  

\subsection{Proposed Network Architecture}
\label{sec:Network}

    \begin{figure*}[!ht]
      \centering
        \includegraphics[width=0.8\linewidth]{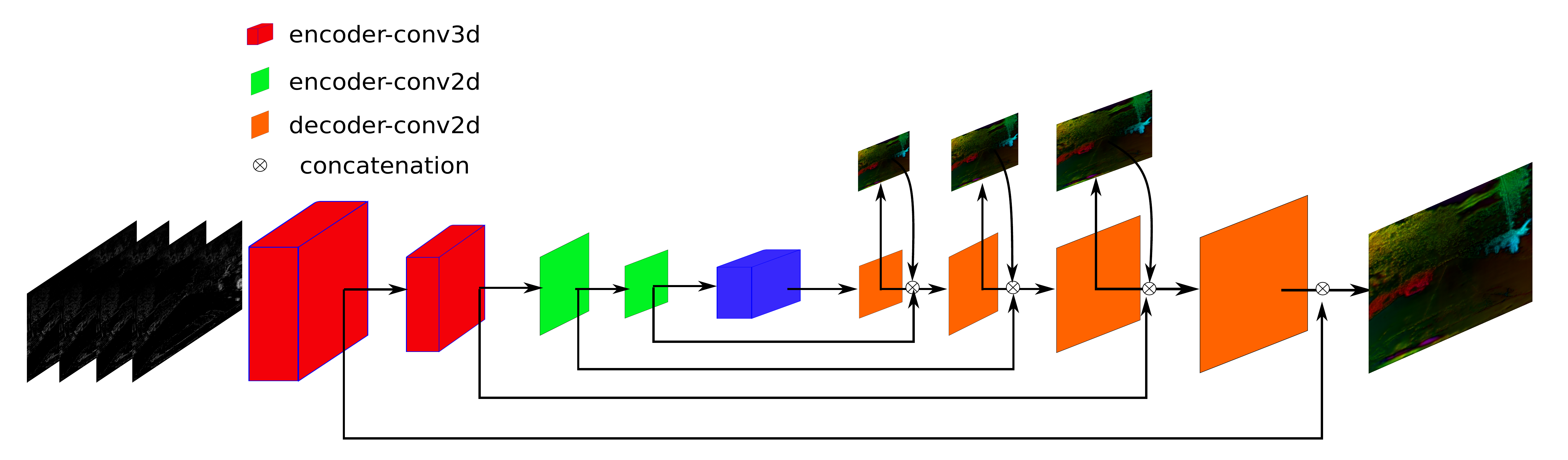}
      \caption{Network structure of the 3D-FlowNet.}
      \label{fig:network-structure}
    \end{figure*}
    
With the input representation \(I_{4,D/2,H,W}\) discussed in \cref{sec:event-encoding}, we propose the 3D-FlowNet architecture to predict the optical flow values. The 3D-FlowNet’s network adopts an encoder-decoder architecture, containing four encoder layers, two residual blocks, and four decoder layers as shown in Fig. \ref{fig:network-structure}. First, the input event image is passed through two 3D-decoders. The 3D-decoders down-sample the time dimension \(d/2\) to 1, and compress the 3D input into 2D \(((4,D/2,H,W)\rightarrow (4,1,H,W)\rightarrow (4,H,W))\). Then the resulting activation are passed through two 2D-decoders, two residual blocks, and four 2D-decoders. For each decoder, the activation is up-sampled using the 2D transposed convolution and then convolved, to obtain the final optical flow estimation. 

There is a skip connection from each encoder to the corresponding decoder. For the skip connection between 2D-encoder and 2D-decoder, the activation of the encoder is directly concatenated with the intermediate optical flow value and the activation of decoder. For the skip connection between 3D-encoder and 2D-decoder, the 3D activation \((C \times D\times W\times H)\) is flattened into 2D tensor \(((C*D)\times W\times H)\) first, then it can be concatenated with the activation of the decoder and the intermediate optical flow. The predicted optical flows are then used together with the grayscale image for the loss calculation.

\subsection{Self-Supervised Loss}
\label{sec:self-supervised-loss}
The event-based camera is a sensor that can produce synchronous grayscale images and asynchronous event data streams simultaneously. Compared to frame-based camera datasets, the number of available event-based camera datasets with annotated labels suitable for optical flow estimation is relatively small. As a result, for training our Spike-FlowNet, we used a self-supervised learning method that uses proxy labels from the recorded grayscale images \cite{jjyu2016unsupflow,UnFlow}.

The total loss consists of a smoothness loss (\(L_{smooth}\)) and a photometric reconstruction loss (\(L_{photo}\)) \cite{jjyu2016unsupflow}. The network needs a pair of grayscale images \((I_t,I_{t+\Delta t})\) to calculate the photometric loss, as well as the event data in the time window \((t,t+\Delta t)\). The second grayscale image is warped to the first grayscale image using the network's predicted optical flow. The photometric loss (\(L_{photo}\)) is used to minimize the difference between the first grayscale image and the inversely warped second grayscale image. This loss is based on the photometric consistency assumption, which states that a pixel value from the first image will be similar to the second frame warped by the predicted optical flow. The photometric loss can be written as:
    \begin{equation}\label{loss}
    \begin{split}
        &L_{loss}(u,v,I_t,I_{t+\Delta t}) = \\
        &\sum_{x,y} \rho (I_t(x,y) - I_{t+dt}(x+u(x,y),y+v(x,y)))
        \end{split}
    \end{equation}
Then, the smoothness loss is adopted to improve the spatial consistency of neighboring optical flow. It is calculated as:
    \begin{equation}\label{smooth}
        \begin{split}
        L_{smooth} = &\sum_{i} \sum_{j} ( \lvert \lvert u_{i,j} - u_{i+1,j}\rvert \rvert +\lvert \lvert (u_{i,j} - u_{i,j+1}\rvert \rvert +\\
        &\lvert \lvert (v_{i,j} - v_{i+1,j}\rvert \rvert +\lvert \lvert (v_{i,j} - v_{i,j+1}\rvert \rvert)
        \end{split}
    \end{equation}

The total loss for the training is computed as the weighted sum of the photometric and smoothness loss:
    \begin{equation}\label{totalloss}
        L_{total} = L_{photo} + \lambda L_{smooth}
    \end{equation}
where \(\lambda \) is the weight factor.

\section{Experiments}

\subsection{Dataset and Implementation Details}

The MVSEC dataset \cite{MVSEC} is used in this paper for training and evaluating the optical flow predictions. The MVSEC dataset contains stereo event-based camera data, including flying, driving, and handheld scenes. Moreover, the dataset provides ground truth poses and depths maps for each event-based camera, and the ground truth optical flow can be generated accordingly. To offer fair comparisons with prior works \cite{Spike-FlowNet,Zhu-RSS-18}, only the outdoor day2 sequence is used for training. 

During the training, the input is centrally cropped to \(256 \times 256\) size. The ADAM optimizer is used, and the initial learning rate of 1e-4. The model is trained for 30 epochs with a batch size of 16, while \cite{Spike-FlowNet} takes 100 epochs. This is because the training of the ANN is faster and more stable than the SNN one. 

\subsection{Results}

\begin{table*}[!ht]
        \caption{Quantitative assessment of our approach compared to EV-FlowNet and Spike-FlowNet}
        \resizebox{\textwidth}{!}{%
        \begin{tabular}{ccccccccc}
        \hline
        \multirow{2}{*}{}   & \multicolumn{2}{l}{outdoor day1} & \multicolumn{2}{l}{indoor flying1} & \multicolumn{2}{l}{indoor flying2} & \multicolumn{2}{l}{indoor flying3} \\\cline{2-9} 
                    & AEE  \(\downarrow\)         & Outlier     \(\downarrow\)      & AEE       \(\downarrow\)      & Outlier     \(\downarrow\)       & AEE         \(\downarrow\)    & Outlier     \(\downarrow\)       & AEE     \(\downarrow\)        & Outlier     \(\downarrow\)       \\ \cline{1-1} \cline{2-9} 
        EV-FlowNet \cite{Zhu-RSS-18}          & 0.49          & 0.2              & 1.03           & 2.2               & 1.72           & 15.1              & 1.53           & 11.9              \\

        Spike-FlowNet \cite{Spike-FlowNet}       &  \(\bm{0.49}\)           &       -           & 0.84           &    -               & 1.28           &        -           & 1.11           &   -                \\
        Ours &      0.51         &   \(\bm{0.1}\)   &    \(\bm{0.7}\)       & \(\bm{0.1}\)       &     \(\bm{1.10}\)  &   \(\bm{0.2}\)&    \(\bm{0.91}\) &    \(\bm{0.1}\)     \\ \hline
        \end{tabular}%
        }
    \label{tab:Quantitative}
    \end{table*}
    \begin{figure*}[!ht]
      \centering
            \includegraphics[width=0.85\linewidth]{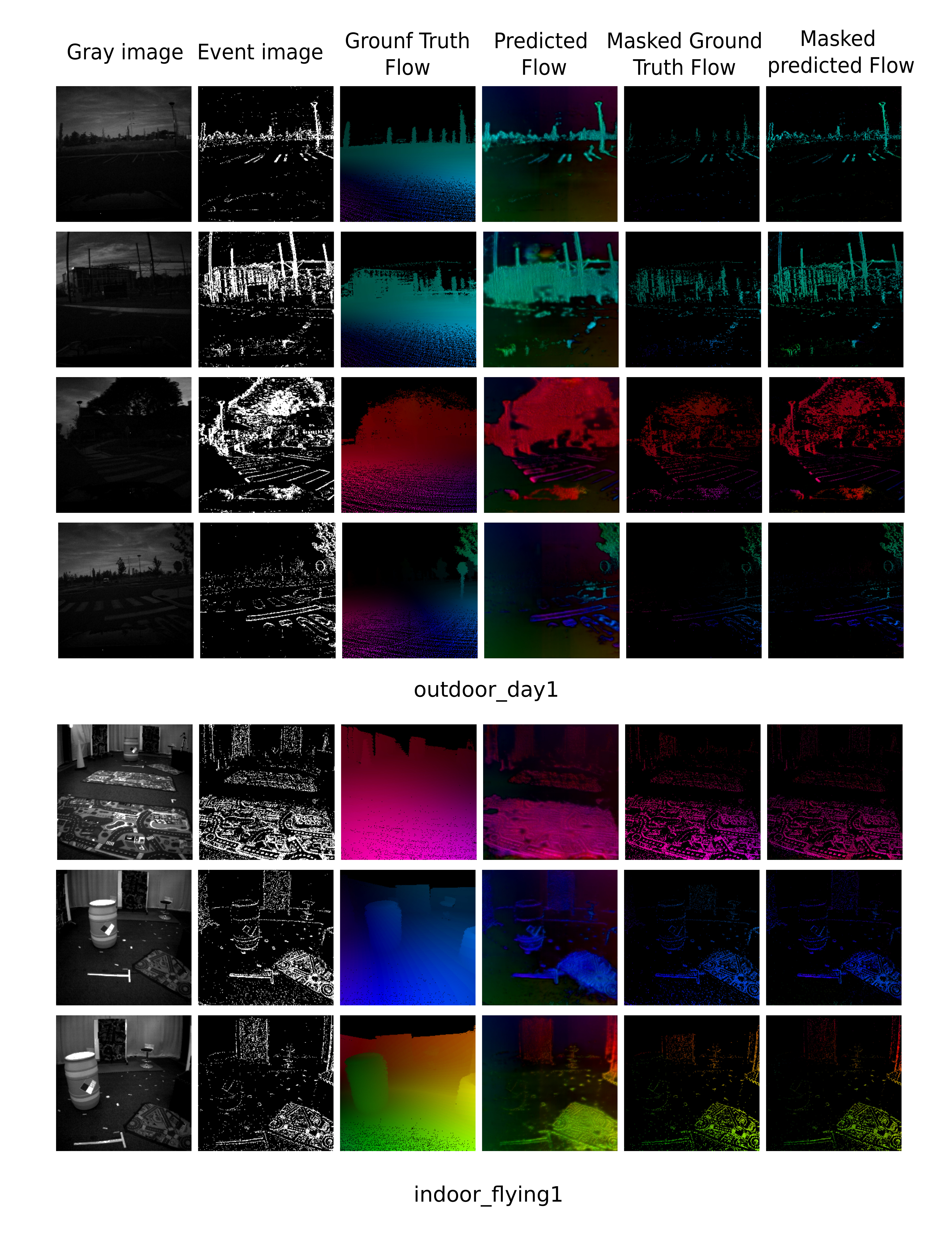}
      \caption{Visualization of the optical flow estimation.}
      \label{fig:res}
    \end{figure*}

Here, the Average End-point Error (AEE) is used to evaluate the optical flow result, and it is defined as:
    \begin{equation}
        AEE = \frac{1}{n}\sum_n \| (u, v)_{pred} - (u, v)_{gt} \|^2
    \label{eqAEE}
    \end{equation}
Where \(n\) is the number of the active pixel in the event image. We also count the outliers that corresponds to the percentage of points with AEE exceeding three pixels. For each sequence, the AEE is calculated in pixels, and the \%Outlier is defined as the percentage of points with \(AEE < 3\) pix. During the testing, the optical flow is also estimated on the centrally cropped $256 \times 256$ event images. The sequences of indoor flying 1,2,3 and outdoor day 1 are used. We use all events from the indoor flying sequences and take events within 800 gray scale frames for the outdoor day1 sequence similar to \cite{Spike-FlowNet}. 

Table \ref{tab:Quantitative} show the results of the AEE evaluation in comparison to previous event-based camera-based optical flow estimation approaches. Our approach achieves better performances than the others in all the indoor\_flying sequences. Our AEE performance is similar to the others in the outdoor\_day1 sequence, but we obtain fewer outliers. Fig. \ref{fig:res} shows the qualitative results of our approach. The grayscale, event image, ground truth flow, and corresponding predicted flow images are displayed in this figure. We mask out the optical flow at points where the event data are absent. The masked optical flow is used here because event-based cameras detect the brightness change at pixels. Low texture regions, such as flat surfaces, produce very few events due to fewer brightness changes, resulting in few optical flow predictions in the corresponding areas. Overall, the results show that 3D-FlowNet can predict optical flow accurately in both indoor and outdoor day1 sequences. This proves that the proposed 3D-FlowNet generalizes well to a variety of environments. 

\section{Conclusion}
In this work, we propose 3D-FlowNet, a deep neural network for optical flow estimations using event-based camera data. We improved the encoding methods for the event data and self-training strategy for the network. The results show that our approach can generate more accurate (13\%~32\%) optical flow estimations. For future work, we hope to combine frame-based cameras with event-based cameras to achieve better and more robust performance in various scenarios.


\end{document}